%% file: not-anonymous-submission-latex-2023.tex
\documentclass[letterpaper]{article} 
\usepackage{aaai23_real}  
\usepackage{times}  
\usepackage{helvet}  
\usepackage{courier}  
\usepackage[hyphens]{url}  
\usepackage{graphicx} 
\usepackage{natbib}  
\usepackage{caption} 
\usepackage{algorithm}
\usepackage{algorithmic}
\usepackage{amsmath}
\usepackage{amsfonts}
\usepackage{booktabs}
\usepackage{bm} 
\usepackage{subcaption}
\usepackage{natbib}
\usepackage{cite}
\usepackage{appendix}
\usepackage{xcolor}         

\usepackage{newfloat}
\usepackage{listings}
\DeclareCaptionStyle{ruled}{labelfont=normalfont,labelsep=colon,strut=off} 
\floatstyle{ruled}
\newfloat{listing}{tb}{lst}{}
\floatname{listing}{Listing}
\title{Solving the Diffusion of Responsibility Problem in \\ 
Multiagent Reinforcement Learning 
with a Policy Resonance Approach}

\author {
    Qingxu Fu       \textsuperscript{\rm 1,2},
    Tenghai Qiu     \textsuperscript{\rm 1,*}, 
    Jianqiang Yi    \textsuperscript{\rm 1,2}, 
    Zhiqiang Pu     \textsuperscript{\rm 1,2}, 
    Xiaolin Ai      \textsuperscript{\rm 1,2},
    Wanmai Yuan     \textsuperscript{\rm 3},
}
\affiliations {
    \textsuperscript{\rm 1} Institute of Automation, Chinese Academy of Sciences, Beijing 100190, China.\\
    \textsuperscript{\rm 2} School of Artificial Intelligence, University of Chinese Academy of Sciences, Beijing 100049, China.\\
    \textsuperscript{\rm 3} CETC Information Science Academy, Beijing 100049, China. \\
    fuqingxu2019@ia.ac.cn, tenghai.qiu@ia.ac.cn, jianqiang.yi@ia.ac.cn, \\
    zhiqiang.pu@ia.ac.cn, xiaolin.ai@ia.ac.cn, Yuanwanmai7@163.com\\
}

\usepackage{bibentry}

\begin{document}

\maketitle

\begin{abstract}
    \input{1-abs-v3.md}
    \end{abstract}
    \input{2-intro.tex}

\input{3-preliminaries.tex}
    \input{4-cause_of_dr.tex}

\input{5-policyrsn.tex}

\input{6-exp_results.tex}

    \section{Conclusions}

\input{7-conclusion.md}

    \bibliography{cite.bib}

\input{8-apendix_real.tex}

\end{document}

%% file: 1-abs-v3.md
SOTA multiagent reinforcement algorithms distinguish themselves in many ways from their single-agent equivalences, except that they still totally inherit the single-agent exploration-exploitation strategy. We report that naively inheriting this strategy from single-agent algorithms causes potential collaboration failures, in which the agents blindly follow mainstream behaviors and reject taking minority responsibility. We named this problem the diffusion of responsibility (DR) as it shares similarities with a same-name social psychology effect. In this work, we start by theoretically analyzing the cause of the DR problem, emphasizing it is not relevant to the reward crafting or the credit assignment problems. We propose a Policy Resonance approach to address the DR problem by modifying the multiagent exploration-exploitation strategy. Next, we show that most SOTA algorithms can equip this approach to promote collaborative agent performance in complex cooperative tasks. Experiments are performed in multiple test benchmark tasks to illustrate the effectiveness of this approach. 

%% file: 2-intro.tex
\section{Introduction}



Recent years have witnessed a growing academic interest in MultiAgent Reinforcement Learning (MARL),
and researchers have proposed a significant number of MARL algorithms to explore the potential of cooperative AIs in many cooperation tasks,
e.g. Starcraft Multi-Agent Challenge (SMAC) \cite{vinyals2017starcraft, usunier2016episodic, lowe2017multi, chen2019crowd, deka2021natural, agarwal2019learning},
Hide-and-Seek with tools \cite{baker2019emergent},
MAgent \cite{rubenstein2014programmable, zheng2018magent, diallo2020multi},
joint-navigation, predator-prey \cite{lowe2017multi} and football \cite{Chenghao2021Celebrating}.
Both policy-based methods \cite{lowe2017multi, iqbal2019actor, yu2021surprising} and 
value-based methods \cite{sunehag2017value, rashid2018qmix, son2019qtran} 
play an essential role in the progress of MARL algorithms.

Nevertheless, the study of MARL is still far from reaching its ambition.
Firstly, 
relatively little research investigates general MARL solutions for large teams with more than 30 agents.
On the one hand, 
simplified grid-based simulation limits the generalization ability of models developed in discrete environments such as MAgents,
On the other hand, 
sophisticated benchmarks such as SMAC (maps allowing up to $27$ agents) are not considering large-scale multiagent teams.
Secondly,
multiagent problems, compared with the single-agent equivalence,
still have underlying issues that have not yet been identified, 
because multiagent interaction grows extremely complicated as the number of agents grows.


This paper reports a newly discovered problem
that surfaces in the multiagent learning process of sophisticated cooperative tasks.
In both value-based and policy-based MARL algorithms,
agents have to explore the environment before they can find better policy or more precise state value estimation,
and to address the exploration-exploitation dilemma \cite{yogeswaran2012reinforcement}.
E.g., value-based algorithms use $\epsilon$-greedy \cite{szita2008many},
and policy-based algorithms have stochastic policies
implemented with entropy regularization that ensures exploration \cite{schulman2017proximal}.
However, 
note that those measures completely inherit from single-agent RL models
without considering any potential problems which are unique to multiagent systems.
In this paper, we reveal that these measures against the exploration-exploitation dilemma during MARL training 
can cause failures in responsibility negotiation between agents.
When a large number of agents are learning to cooperate,
the responsibility of carrying out essential minority behaviors 
(namely minority responsibility) tends to be neglected by ALL agents.
Furthermore, as the number of agents $N$ increases, 
this problem deteriorates and becomes even more significant.

We use a term from the social psychology domain,
Diffusion of Responsibility (DR) \cite{darley1968bystander},
to name this effect because they share similarities in many ways.
In social psychology, 
the diffusion of responsibility effect is also known as the bystander effect.
It attempts to explain a crowd's reaction in situations
where help is needed by one (or a few) bystander(s),
e.g., a pedestrian fainting in the street.
The DR problem shows that, in general,
the willingness of individuals to provide help decreases with the number of bystanders.
Psychologists illustrate that all bystander agents struggle
between taking responsibility and staying out of trouble.
When only one agent is present, 
it takes full pressure of the responsibility to intervene.
However, when multiple agents are present, 
no agents prefer to intervene because the responsibility is diffused among agents, 
then overwhelmed by the motivation of staying away from troubles.

Despite the superficial similarities, 
the underlying mechanism of the DR problem in MARL is different from that in the psychology domain.
We emphasize that
the mechanism of the DR problem is rooted in the exploration-exploitation dilemma 
\textbf{instead of} agents' egoist or selfishness.
The MARL DR problem exists even if agents are rewarded as a team (instead of individuals),
or reward decomposition methods such as Qmix \cite{rashid2018qmix, hu2021rethinking} are adopted.
Alternatively speaking,
we stress that the MARL DR problem is irrelevant to the way the reward is given (team reward or individual rewards), 
nor the way the credit is assigned (with or without value decomposition techniques).

The main contributions of this paper are:
\begin{itemize}
\item We theoretically analyze the feature and the cause of the DR effect 
    in FME, a multiagent environment to test basic cooperative behaviors.
\item We carry out experiments with various algorithms, 
    including Fine-tuned Qmix \cite{hu2021rethinking}, Conc4Hist \cite{fu2022concentration}, and PPO-MA \cite{yu2021surprising},
    to illustrate the nature of the DR problem, which are consistent with the theoretical analysis.
\item We propose a Policy Resonance (PR) approach to resolve the DR problem 
    under the Centralized Training with Decentralized Execution (CTDE) paradigm.
    Policy Resonance is \textbf{not} an independent algorithm,
    instead, it can be flexibly implemented on most existing SOTA host algorithms.
\item We implement our PR approach on algorithms, including PPO-MA and 
    Conc4Hist to demonstrate the effectiveness of our method 
    on multiple benchmarks, including FME and ADCA (a complex swarm combat environment) \cite{fu2022concentration}.

\end{itemize}

%% file: 3-preliminaries.tex
\section{Preliminaries}
\subsection{Exploration-exploitation Dilemma.} \label{sec:explorationDilemma}
An RL algorithm must strike a balance 
between exploration and exploitation in order to train agents effectively \cite{sutton2018reinforcement}.
The exploration-exploitation dilemma has been intensively studied in single-agent problems
for both value-based models and policy-based models.
For value-based models, exploration solutions are $\epsilon$-greedy \cite{szita2008many, sutton2018reinforcement},
Boltzmann exploration \cite{carmel1999exploration, morihiro2006emergence}, probability matching \cite{koulouriotis2008reinforcement} et.al.
And the $\epsilon$-greedy is the most common strategy adopted by most MARL algorithms such as QMIX and Fine-tuned Qmix.
On the other hand,
policy-based models \cite{schulman2017proximal, iqbal2019actor, yu2021surprising} rely on stochastic policies and 
entropy regularization to ensure sufficient exploration \cite{williams1992simple, mnih2016asynchronous}.
In multiagent cases, entropy can be calculated on all agents by:
\begin{equation}\label{eq:entropy}
    H  =  \frac{c_H}{N} \sum_{i=1}^{N}  \mathbb{E}_{u \sim \pi_i(u | o_i)}[-\log \pi_i(u | o_i)]
\end{equation}
where $c_H$ is a coefficient, $\mathcal{A}$ is the set of agents, $N=|\mathcal{A}|$ is the number of agents, 
$\pi_i$ represents agent policies,
$o_i$ represents agent observation and
$\mathcal{U}$ is the set of agent actions, $u \in \mathcal{U}$.

In this paper,
we mainly focus on policy-based models.
When analyzing the cause of DR, 
we assume entropy regularization is used because it is adopted by most SOTA policy-based algorithms.
Additionally, we assume $\epsilon$-greedy is used for the analysis of value-based models in Appendix \ref{Value-Based-Perspective}.

\subsection{Benchmark Environments.}\label{BenchmarkEnvironments}
%



\subsubsection{Fast Multiagent Evaluation (FME).}
To help analyze the Diffusion of Responsibility problem,
we designed a diagnostic environment named Fast Multiagent Evaluation (FME).

FME is designed as a test environment that 
uses \textbf{independent and sequentially arranged levels}
to identify the weakness of given MARL algorithms.
Each level is an abstract mini-task, e.g., multiagent multi-armed bandit problem \cite{sutton2018reinforcement}.
Such design enables us to theoretically analyze issues that surface in experiments.

FME levels only provide \textbf{team reward} to increase difficulty and
to fully expose underlying issues of MARL algorithms.
Moreover, by only using team reward FME encourages agents to behave cooperatively rather than selfishly.

The following levels are needed to analyze the DR issue,
each of them requires only \textbf{single-step} decision:

\textbf{Lv-A}: Finding a static action with reward.
Agents need to find a predefined action $u^\alpha\in \mathcal{U}$ that leads to reward.
Reward $R_A=N_{\mathcal{A}}(u^\alpha)/N$,
where $N_{\mathcal{A}}(u^\alpha)$ counts the number of agents selecting $u^\alpha$. 
The optimal joint policy is for all agents to simply choose $u^\alpha$, 
and the optimal reward expectation is $\mathbb{E}\left(R_A|\boldsymbol{\pi}^*\right)=1$, 
where $\boldsymbol{\pi}^*$ is the optimal joint action.

\textbf{Lv-B}: Multiagent variant of the \textbf{multi-armed bandit} problem.
After every agent has selected an arm on the bandit,
a random arm $u^\beta \in \mathcal{U}$ is sampled from a predefined constant distribution $p_\beta(u)$.
Reward is $R_B=N_{\mathcal{A}}(u^\beta)c_\beta/N$,
where $c_\beta$ is a constant and $N_{\mathcal{A}}(u^\beta)$ counts the number of agents choosing the $u^\beta$ arm.
The optimal solution (proof in Appendix \ref{lvb-reward}) is all agents selecting
the arm with highest probability according to $p_\beta(u)$.
To scale optimal reward expectation $\mathbb{E}\left(R_B|p_\beta(u), \boldsymbol{\pi}^*\right)$ at 1,
the constant $c_\beta=1/{\underset{u}{\mathrm{max}}[p_\beta(u)]}$.

\textbf{Lv-C}: Responsibility assignment challenge.
Agents are rewarded for selecting a predefined constant action $u^\gamma$ which has a maximum capacity $N-1$.
If all $N$ agents choose $u^\gamma \in \mathcal{U}$,
then the capacity of $u^\gamma$ exceeds and reward turns zero. \\
(Informally, $u^\gamma$ can be seen as a fragile bridge that collapses under exceeding weight.
Most agents are responsible for crossing the bridge to reach the destination,
but a minority of agents are responsible for choosing otherwise for the overall interests of the team.)

Formally, the Lv-C reward is:
\begin{equation}\label{lvc-reward}
    R_C= \begin{cases}{N_{\mathcal{A}}\left(u_{\gamma}\right)}/{(N-1)}, 
        & 0\le N_{\mathcal{A}}\left(u_{\gamma}\right)<N \\ 0, 
        & N_{\mathcal{A}}\left(u_{\gamma}\right)=N\end{cases}
\end{equation}
where $N_{\mathcal{A}}(u^\gamma)$ is the number of agents selecting constant action $u^\gamma$. 
The optimal reward expectation is $\mathbb{E}\left[R_C|\boldsymbol{\pi}^*\right]={N_{\mathcal{A}}(u^\gamma)}/{(N-1)}=1$
when $N_{\mathcal{A}}\left(u_{\gamma}\right)=N-1$.

Other properties of FME include:
\begin{itemize}
    \item All levels share the action space $\mathcal{U}$, but each level interprets $\mathcal{U}$ differently.
    \item Agents only observe the level indices and one-hot agent identities, which is already sufficient for optimizing policies.
    \item By configuring predefined parameters such as $u^\alpha$, $p_\beta(\cdot)$, $u^\gamma$ differently, 
    each type of level can be used multiple time in a task setting. 
    \item We use a tag such as \textit{1A4B2C} to describe a FME task. 
    E.g, tag \textit{1A4B2C} represents a FME task composed by one Lv-A, four Lv-B and two Lv-C levels with different configurations.
    \textbf{Detailed parameters} are displayed in Appendix \ref{appendix:FMEtable}.
    
\end{itemize}

\begin{table}[!t]
    \vspace{-0.5cm}

    \centering
    \setlength{\tabcolsep}{0.5mm}{
        \begin{tabular}{ccccccccc}
        \toprule
        \tiny{          }   & \tiny{$N$=3}    & \tiny{$N$=5}     & \tiny{$N$=6}    & \tiny{$N$=7}    & \tiny{$N$=8}      & \tiny{$N$=15}    & \tiny{$N$=25}      & \tiny{$N$=40}         \\
        \midrule
        \tiny{PPO-MA         } & \tiny{1.00} & \tiny{1.00}  & \tiny{0.87} & \tiny{0.00} & \tiny{0.00}  & \tiny{0.00}   & \tiny{0.00}   & \tiny{0.00}      \\
        \tiny{Fine-tuned Qmix} & \tiny{1.00} & \tiny{1.00}  & \tiny{1.00} & \tiny{1.00} & \tiny{0.95}  & \tiny{0.85}   & \tiny{0.27}   & \tiny{0.16}   \\
        \bottomrule
    \end{tabular}
    }
    \vspace{-0.2cm}
    \caption{
        The average rewards $R_C$ in Lv-C evaluation.
        Trained and tested in FME diagnostic environment with setting \textit{1A4B2C}.
        (only Lv-C reward is displayed, optimal reward of Lv-C is 1). 
    }
    \label{lvc-problem-intro}
    \vspace{-0.4cm}
\end{table}

\subsubsection{ADCA.}
The Asymmetric Decentralized Collective Assault (ADCA)
benchmark is a swarm combat environment modified and 
improved from \cite{fu2022concentration}.
As shown in Fig.\ref{fig:TwoStagePR-ADCA}, in ADCA, two asymmetric teams compete with each other.
The RL-controlled team has $N=50$ agents,
the enemy team that is controlled by script AI.
The enemy team has more agents ($N' \in [120\%N, 150\%N]$) and weapon advantage,
see Appendix \ref{appendix:adca} for ADCA details.

\subsection{Base Algorithms.}
Following algorithms are used in analysis and experiments:
\begin{itemize}
    \item Fine-tuned Qmix by \citeauthor{hu2021rethinking}. Value-based algorithm.
    \item PPO-MA, similar to the model proposed by \citeauthor{yu2021surprising}, but with action masking and input shaping tricks removed. Policy-based algorithm.
    \item Conc4Hist by \citeauthor{fu2022concentration}. Policy-based algorithm.
\end{itemize}

%% file: 4-cause_of_dr.tex
\section{Brief Analysis of the DR Problem}\label{sec:Analysis}

Without appropriate management,
a team with abundant agents can fail to complete simple tasks 
that could have been done by just a few agents.
In this work, 
we illustrate the DR problem using \textit{FME's responsibility assignment challenge (Lv-C) in Section \ref{BenchmarkEnvironments}}.
The ideal joint policy is for most team agents to choose the $u^\gamma$ action,
while a minority of agents choose actions other than $u^\gamma$.
And the optimal policy is for all but one agent to constantly choose $u^\gamma$.
A misleading trap policy is that all agents become addicted to action $u^\gamma$ 
and eventually exceeding the limited capacity.


\begin{figure}[!t]
    \vspace{-0.65cm}
    \centering
    \includegraphics[width=\linewidth]{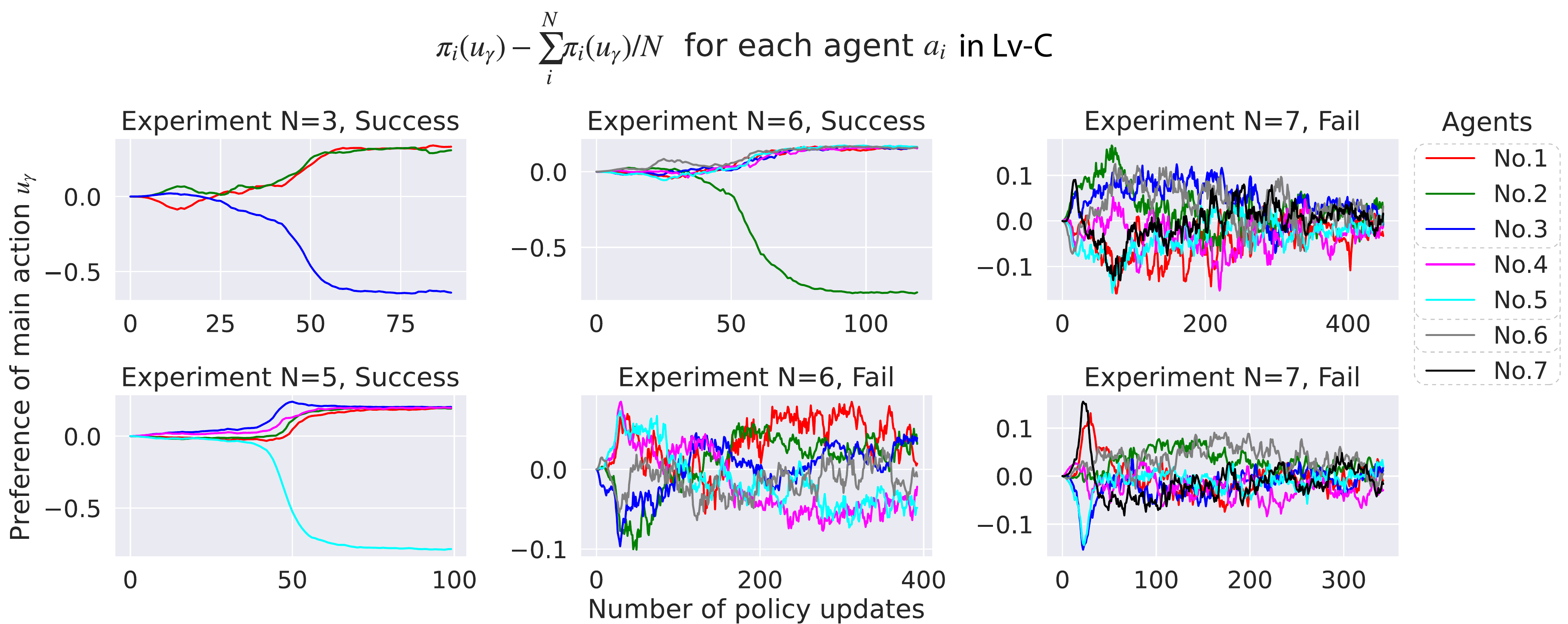}
    \caption{
        Six individual experiments from Table.\ref{lvc-problem-intro},
        showing the preference development toward $u^\gamma$ of the individual agent in 
        in an FME Lv-C (last Lv-C in \textit{1A4B2C} sequence).
        Using $\pi_i(u_{\gamma})-\frac{1}{N}\sum_{j \in \mathcal{A}}\pi_j(u_{\gamma})$
        to describe the preference of each agent compared with the team average.
    }
    \vspace{-0.4cm}
    \label{fig:PersonalityDevelop}
    \end{figure}

We train agents in FME diagnostic environment with sequential setting \textit{1A4B2C} with
PPO-MA (representing policy-based algorithm) and Fine-tuned Qmix (representing value-based algorithm), respectively.
Results in Table.\ref{lvc-problem-intro} show that 
PPO-MA is misled into the capacity-exceeding trap when the $N \ge 6$.
Fine-tuned Qmix survives more agents but still fails when $N \ge 15$.

Focusing on \textbf{policy-based} PPO-MA algorithm,
we sample six individual experiments within $3 \le N \le7$ 
and display agents' policy development process in Fig.\ref{fig:PersonalityDevelop}.
To investigate how agents negotiate responsibility about $u^\gamma$,
we use $f(\pi_i) = \pi_i(u_{\gamma})-\frac{1}{N}\sum_{j \in \mathcal{A}}\pi_j(u_{\gamma})$
to describe agents' preference towards $u^\gamma$.
Fig.\ref{fig:PersonalityDevelop} shows that, when $N \le 5$,
agents successfully negotiate on a optimal responsibility assignment plan 
by making \textit{one of the team agents to give up its $u^\gamma$ preference}.
However, when $N=6$ the negotiation fails randomly when the experiment is repeated.
And $N \ge 7$ makes the cooperation impossible.

Next we briefly analyze the reason that policy-based PPO-MA falls into trap in this task:

\textbf{1.} At the beginning of the training process,
the agents are initialized with random and high-entropy policies,
preferring each action (including $u^\gamma$) almost equally \cite{saxe2013exact, mcelreath2020statistical}.
Let $\varPi_0 = \underset{i}{\operatorname{max}} \pi_i (u_{\gamma}|o_{i} )$ 
be the max probability (among agents) of selecting $u^\gamma$.
Since the initial policy is random, then $\varPi_0 \approx \frac{1}{n_k}$,
where $n_k = |\mathcal{U}|$ is the size of action space.

\textbf{2.} Therefore, at the beginning
the capacity-exceeding penalty has a very low triggering probability $P_{plt}$:
\begin{equation}\label{init}
P_{plt} = P\left[N_{\mathcal{A}}\left(u_{\gamma}\right)=N |\boldsymbol{\pi} \right] 
        = \prod_{i=1}^{N} \pi_i (u_{\gamma}|o_{i} ) 
        \le \varPi_0^N
\end{equation}
e.g., when $n_k=10$, $N=15$ and then $P_{plt}\approx 10^{-15}$, 
meaning that only one capacity-exceeding sample is expected within $10^{15}$ samples.
Unless the number of agents $N$ is small enough,
the learning algorithm cannot notice the capacity-exceeding trap at the beginning.
From another perspective,
the learning algorithm also \textbf{cannot distinguish an Lv-C from an Lv-A at the initial stage}
due to the lack of samples with the capacity-exceeding penalty.
Therefore, the algorithm will encourage all agents to select $u_{\gamma}$,
increasing $\pi_i (u_{\gamma})$ gradually and synchronously $\forall i\in \mathcal{A}$.

\textbf{3.} 
Then we prove that when $N$ is large enough,
\textbf{agents cannot stop the increase of $\pi_i (u_{\gamma})$ until suppressed by entropy regularization}.
Firstly, find the necessary conditions for $\pi_i (u_{\gamma})$ to stop growing.
For clarity, we shorten 
$N_{\mathcal{A}}\left(u_{\gamma}\right)=N_{\gamma}$ and shorten $\pi_i (u_{\gamma}|o_i)=\pi_{i\gamma}$.
For an agent $i$, the growth of $\pi_{i\gamma}$ will be stopped and reversed 
when larger $\pi_{i\gamma}$ benefits the team reward no more:
\begin{equation}\label{eq:partial}
    \begin{split}
        \frac{\partial \mathbb{E}[R_C(\boldsymbol{\pi})] }{\partial \pi_{i\gamma}} < 0
    \end{split}
\end{equation}
And from Eq.\ref{lvc-reward}:
\begin{equation}\label{eq:new_exp}
\begin{split}
    \mathbb{E}[R_C(\boldsymbol{\pi})] 
    &= \sum_{k=0}^{N-1} \frac{k}{N-1} P[N_{\gamma}=k] + 0 \cdot P[N_{\gamma}=N] \\
    &= \sum_{k=0}^{N} \frac{k}{N-1} P[N_{\gamma}=k] - \frac{N}{N-1} P[N_{\gamma}=N] \\
    &= \frac{1}{N-1} \sum_{j=1}^{N} \pi_{j\gamma} - \frac{N}{N-1} \prod_{j=1}^{N} \pi_{j\gamma}
\end{split}
\end{equation}
From Eq.\ref{eq:new_exp} and Eq.\ref{eq:partial}:
\begin{equation}\label{eq:xx0}
    \begin{split}
        \frac{\partial \mathbb{E}[R_C(\boldsymbol{\pi})] }{\partial \pi_{i\gamma}} = 
        \frac{1}{N-1} - \frac{N}{N-1} \prod_{j\in[1,N], j\neq i} \pi_{j\gamma} < 0
    \end{split}
\end{equation}
\begin{equation}\label{eq:xx1}
    \begin{split}
        \sum_{j\in[1,N], j\neq i} \log \pi_{j\gamma} > -\log N
    \end{split}
\end{equation}
Define a probability $\varPi_{i\gamma}$ to simplify Eq.\ref{eq:xx1}, s.t.:
\begin{equation}\label{eq:xx2-1}
    \log\varPi_{i\gamma}=\frac{1}{N-1}\sum_{j\in[1,N], j\neq i} \log \pi_{j\gamma}
\end{equation}
From Eq.\ref{eq:xx1} and Eq.\ref{eq:xx2-1} we have:
\begin{equation}\label{eq:xx2-2}
\begin{split}
    \varPi_{i\gamma} > N^{-\frac{1}{N-1}}
\end{split}
\end{equation}
Because of Eq.\ref{eq:xx2-1} and the monotonicity of the log function,
there exists at least one agent $m\in [1,N], m \neq i$ that satisfies $\pi_{m\gamma} \ge \varPi_{i\gamma}$,
therefore:
\begin{equation}\label{eq:xx3}
\begin{split}
    \pi_{m\gamma}  > N^{-\frac{1}{N-1}}
\end{split}
\end{equation}
Unfortunately, as $N$ increase, Eq.\ref{eq:xx3} becomes impossible to be satisfied:
\begin{itemize}
    \item RHS increases and approaches 1, $\lim_{N \to \infty} N^{-\frac{1}{N-1}} = 1$. E.g., when $N=15$, RHS = 0.824; $N=50$, RHS = 0.923.
    \item LHS is limited by entropy regularization in Eq.\ref{eq:entropy}, Section \ref{sec:explorationDilemma} assumption.
    In this case, entropy loss prevents $\pi_{j\gamma}$ from 
    getting closer to 1 (policy becoming deterministic) for all agents.
\end{itemize}
Therefore, we can always find $N_{max}$ such that if $N>N_{max}$,
then Eq.\ref{eq:xx3} cannot be satisfied (by any agent),
consequently, Eq.\ref{eq:partial} cannot be satisfied.
Therefore, 
agents $i$ will race each other to increase the probability $\pi_{i\gamma}$ 
until $\pi_{i\gamma}$ is suppressed by entropy loss,
which explains the failed experiments in Fig.\ref{fig:PersonalityDevelop}.

This result is counterintuitive, 
it is irreconcilable with our single-agent RL knowledge because this is a phenomenon \textbf{unique to multiagent} RL algorithms.
Recall that we are using team rewards to train agents,
the environment only encourages agents to act cooperatively instead of selfishly.
However,
while any agent can greatly benefit from switching their responsibility 
from majority action $u^{\gamma}$ to minority actions $u^{-\gamma}$,
none of them will actually do that due to this Diffusion of Responsibility (DR) problem.

Now we have reached the core of the DR problem.
We cannot simply blame the DR problem solely on entropy regularization and
believe the DR problem exists only in policy-based algorithms.
In fact,
\begin{itemize}
    \item The DR problem also exists when \textbf{value-based} algorithms are applied. (See Table.\ref{lvc-problem-intro} and Fig.\ref{fig:DRWithMoreAgents} Fine-tuned Qmix).
    \item The DR problem originates from \textbf{$\epsilon$-greedy} in the case of value-based algorithms. (See Appendix \ref{Value-Based-Perspective}).
    \item Entropy regularization and $\epsilon$-greedy are both measures to solve the exploration-exploitation dilemma.
\end{itemize}
With these clues,
we can propose a much more intuitive and general explanation:

The exploration behavior of agents has a significant influence on each other in MARL,
but concurrent MARL learners cannot correctly perceive this influence.
From the perspective of a MARL learner, 
this mutual influence caused by \textbf{internal} exploration is mistakenly perceived as
\textbf{external} environment characteristics.
This bias of perception eventually leads to failure in responsibility negotiation.
E.g.,
MARL learners \textbf{cannot correctly distinguish an Lv-C from an Lv-B (multi-arm bandit)} in FME,
because the learners mistakenly regard the Lv-C $u^{\gamma}$ action as a bandit arm that fails randomly,
not knowing that the randomness is actually caused by the \textbf{internal} exploration.
In conclusion,
the multiagent exploration strategy is the root of the DR problem,
and it is more reasonable to trace the cause of the DR problem down to the exploration-exploitation dilemma.


\section{The Non-triviality of the DR Problem}
In previous sections,
we theoretically analyze the DR problem in FME diagnostic environment,
but a few important questions are left unanswered:
\begin{enumerate}
    \item \textit{Is DR a general problem 
    that can disguise and merge itself into sophisticated MARL tasks 
    such as Starcraft, ADCA, et al.,
    or is it simply a phenomenon that exists only in our diagnostic environment?}
    \item \textit{Can existing algorithms solve the DR problem?}
    \item \textit{For a black box MARL task, do we need to explicitly identify DR problems before dealing with them?}
\end{enumerate}

\textbf{1.} The DR problem is a general issue in MARL.
The FME diagnostic environment is an abstraction and reflection of actual scenarios 
agents encounter in complicated cooperation tasks such as SMAC, ADCA, etc.
The DR problem in FME seems easy to solve via reward hand-crafting,
but such a solution cannot handle DR issues 
hidden behind the scenarios that FME reflected.

For instance,
since the observation area is limited in SMAC and ADCA,
a minority of agents need to patrol unknown area
while the majority of them deal with already known enemies.
Furthermore,
RL algorithms make decisions periodically instead of constantly in SMAC.
If an enemy $X$ is killed halfway through a decision step,
agents with the \textit{Attack-$X$} command become idle,
even if there is another enemy $Y$ firing at them.
In this case,
the best policy is for the majority of agents to instantly turn to $Y$
when the HP of $X$ is low, 
leaving only a minority of agents to finish off $X$.
Moreover, similar scenarios also exist in ADCA.
When a group of agents is being pursued by enemies with numerical superiority,
the majority of agents need to retreat and regroup somewhere with terrain advantage,
while a minority of agents have to stay behind to slow down pursuers 
(by eliminating several frontline enemies).
When the number of agents $N$ is large,
all those scenarios can trigger the DR problem that we have seen in FME Lv-C.

\textbf{2.}
There are no concurrent and non-task-specific methods against the DR problem.
It is difficult, if not impossible, 
to apply methods such as reward-crafting to these sophisticated scenarios described above.
Furthermore,
there may be forms of cooperation actions
beyond our cognition in large multiagent systems, 
e.g., some cooperative behaviors of swarming insects.
If a multiagent system grows large enough,
MARL algorithms will inevitably produce cooperative behaviors beyond our understanding.
If DR problems occur in such systems,
we can only solve them with \textbf{general and non-task-specific} methods.

\textbf{3.}
As this section has revealed, 
it is difficult to analyze DR problems rigorously in multiagent systems.
However,
\textbf{if a general approach can be found to solve the DR problem 
without damaging the original algorithm structure and performance,
we can safely apply this approach without even knowing the existence of the DR problem.}
The next section proposes a Policy Resonance approach to achieve this objective.

%% file: 5-policyrsn.tex
\section{Policy Resonance}

\begin{figure}[!t]
    \vspace{-1cm}
        \centering
        \includegraphics[width=0.8\linewidth]{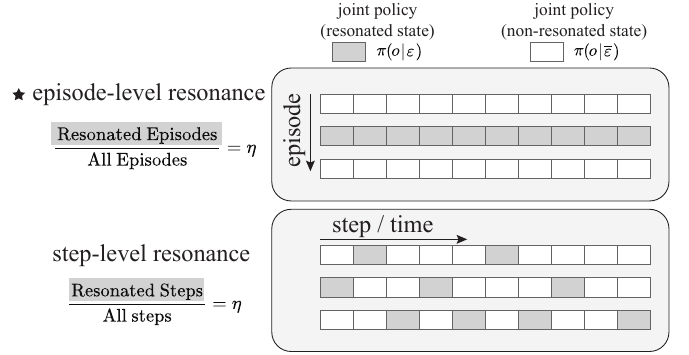}
    \vspace{-0.1cm}
    \caption{The policy resonance procedure.}
        \label{fig:PolicyResonance}
    \vspace{-0.5cm}
    \end{figure}
    
\begin{figure*}[!t]
    \vspace{-0.7cm}
    \centering
    \includegraphics[width=0.8\linewidth]{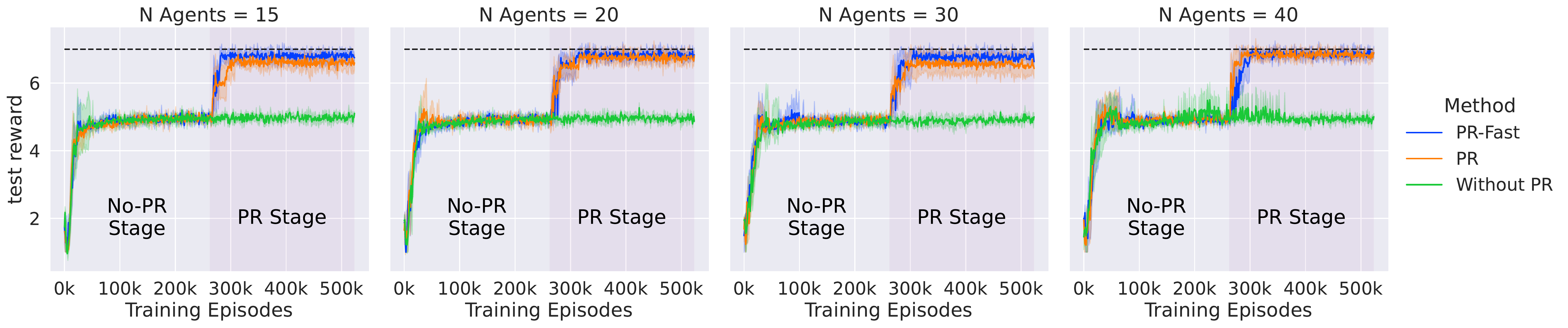}
    \vspace{-0.1cm}
    \caption{
        Two-stage training using Policy Resonance and PPO-MA on FME with setting \textit{1A4B2C}.
        Policy Resonance (PR) is used in the second stage.
    }
    \label{fig:TwoStagePR-FME}
    \vspace{-0.3cm}
  \end{figure*}
\begin{figure*}[!h]
    \centering
    \includegraphics[width=0.8\linewidth]{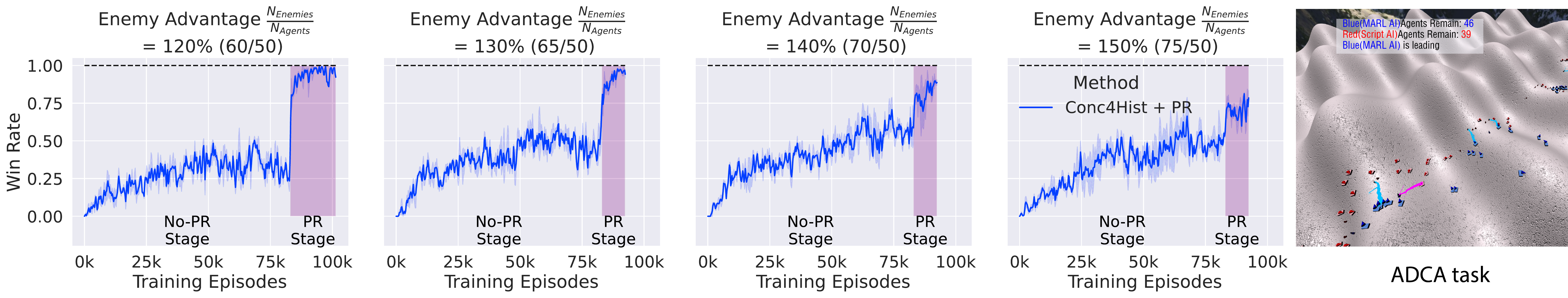}
    \caption{
Training agents to cooperate in the ADCA environment using Conc4Hist implemented with PR. 
Policy Resonance (PR) is used in the second stage.
    }
    \vspace{-0.5cm}
    \label{fig:TwoStagePR-ADCA}
\end{figure*}

To key to addressing the DR problem is to find a better solution to deal with the multiagent version of the exploration-exploitation dilemma,
and eventually, we expect agents to diverge their behaviors and take explicit and well-divided responsibilities spontaneously.

The method we propose is Policy Resonance,
which is designed under the policy-based algorithm framework with the following features:
\begin{itemize}
    \item The motivation is to modify the joint exploration behavior,
    so as to improve concurrent solutions against the exploration-exploitation dilemma.
    \item Not task-specific, applicable to all MARL tasks.
    \item Designed \textbf{as a plugin} module. Not independent. Removable once training is done.
    \item Strictly \textbf{following the CTDE paradigm}. 
\end{itemize}

In existing SOTA MARL models, 
an agent, without negotiating with other agents, has \textbf{full autonomy} to decide whether it should explore at any moment.
This practice is a convention directly inherited from the single-agent RL algorithms,
as in a single-agent RL task an agent has no one to cooperate with.
According to our DR problem analysis in Section \ref{sec:Analysis},
if we deliberately make agents agree tacitly on the opportunity to carry out exploration or exploitation,
the mutual exploration influence can be resolved.
Note that since exploration happens only in the centralized training (CT) stage,
we are not breaking the CTDE paradigm.

We use \textbf{policy resonance} to describe this tacit agreement on 
exploration-exploitation behaviors within a multiagent team.
As a plugin implementation,
our approach extends from the raw policy of the host RL model,
we use $\boldsymbol{\pi}=\lbrace \pi_1, \pi_2, \dots \rbrace$ to represent the raw joint policy,
where $\pi_i$ represents the raw individual policy of agent $i$.
After policy resonance is enabled,
$\boldsymbol{\pi}$ will be mapped into a carefully designed joint policy $\boldsymbol{\pi}'$
which considers agent-wise synchronization on exploration-exploitation behaviors.
$\boldsymbol{\pi}' = \lbrace \pi_1', \pi_2', \dots \rbrace$.

When policy resonance is enabled, 
the new joint policy $\boldsymbol{\pi}$ will constantly and randomly switch between two states:
\textbf{resonated state} $\boldsymbol{\pi}'[\mathbf{u} | \varepsilon]$ and \textbf{non-resonated state} $\boldsymbol{\pi}'[\mathbf{u} | \overline{\varepsilon} ] $.
The selection of states is controlled by a probability $\eta$ named \textbf{resonance probability},
the probability of choosing the resonated state is $P(\varepsilon)=\eta$, and $P(\overline{\varepsilon})=1-\eta$, and:
\begin{equation}
    \label{eq:resonated}
        \boldsymbol{\pi}'[\mathbf{u}] = 
        P(\varepsilon)            \boldsymbol{\pi}'[\mathbf{u} | \varepsilon]  + 
        P(\overline{\varepsilon}) \boldsymbol{\pi}'[\mathbf{u} | \overline{\varepsilon} ] 
\end{equation}
The change of resonating state also influences each individual policy $\pi_i$:
\begin{equation}
    \label{eq:resonated-idv}
        {\pi}_i'({u_i}) = 
        P(\varepsilon)            {\pi}_i'({u_i} | \varepsilon)  + 
        P(\overline{\varepsilon}) {\pi}_i'({u_i} | \overline{\varepsilon} ) 
\end{equation}
\textbf{In non-resonated state}, agents make decisions autonomously and independently with individual policies $\lbrace \pi_1', \pi_2', \dots \rbrace$:
\begin{equation}\label{eq:non-resonated}
    \boldsymbol{\pi}'[\mathbf{u} | \overline{\varepsilon}]=\prod_{i=1}^{N} \pi_{i}^{\prime} ( u_{i} | \overline{\varepsilon})
\end{equation}
where $\pi_{i}^{\prime} ( u_{i} | \overline{\varepsilon})$ will be determined later.\\
\textbf{In resonated state}, agents will synchronously avoid exploration by exploiting current policies greedily:
\begin{equation}
    \label{resonatedindividual}
    \pi^{\prime}_i (u_i |   \varepsilon )=\left\{\begin{array}{ll}
        1, & u_i=u_i^* \\
        0, & \text{otherwise}
    \end{array}\right.
\end{equation}
\begin{equation}
\begin{split}
    \boldsymbol{\pi}'[\mathbf{u} |  \varepsilon]&=\left\{
        \begin{array}{ll}
            1, & \mathbf{u}=\mathbf{u}^{*} \\
            0, & \text{otherwise}
        \end{array}
        \right.
\end{split}
\end{equation}
where $\mathbf{u}^{*} = \lbrace u_1^*, u_2^*, \dots \rbrace$ is joint greedy action, 
and $u_i^* = {\operatorname{argmax}}_{u\in\mathcal{U}}[\pi_i(u)]$ is the most prefered action 
by current raw individual policy $\pi_i$.

Finally, 
we need to determine the mapping from raw individual policy $\pi_i$ to ${\pi}_i'({u_i} | \overline{\varepsilon})$.
Although policy resonance has done lots of modification on raw joint policy $\boldsymbol{\pi}$,
we can still \textbf{make individual policy invariant}, i.e. $\pi^{\prime}_i (u_i) = \pi_i (u_i)$.
In this way, 
this policy resonance process becomes completely transparent to host algorithms,
and will not damage the overall algorithm structure.
Combining  $\pi^{\prime}_i (u_i) = \pi_i (u_i)$ with Eq.\ref{eq:resonated-idv}:
\begin{equation}\label{eq:yyy1}
    \pi_i (u_i) = 
    \eta      {\pi}_i'({u_i} | \varepsilon)  + 
    (1-\eta)  {\pi}_i'({u_i} | \overline{\varepsilon} ) 
\end{equation}
Then from Eq.\ref{eq:yyy1} and Eq.\ref{resonatedindividual}:
\begin{equation}
    \label{eq:ff1}
    \pi^{\prime}_i (u_i |  \overline{\varepsilon}) = \left\{\begin{array}{ll}
        \frac{\pi_i(u_i^*)-\eta}{1-\eta}, & u_i=u_i^* \\
        \pi_i(u_i )\left[\frac{1-\pi^{\prime}_i \left(u_i^* |   \overline{\varepsilon}\right)}{1-\pi_i(u^*_i)}\right], & \text{otherwise}
    \end{array}\right.
\end{equation}
where $\frac{\pi_i(u_i^*)-\eta}{1-\eta}$ is abbreviated to $\pi^{\prime}_i \left(u_i^* |\overline{\varepsilon} \right)$ when $u_i \neq u_i^*$.

There is a constraint for $\eta$ from Eq.\ref{eq:ff1}.
Since $\pi^{\prime}_i(u_i |  \overline{\varepsilon}) \ge 0$, 
then $\eta$ needs to satisfy $\eta \le \pi_i(u_i^*)$.
E.g., when choosing between $n_k$ discrete actions, 
the minimum possible value of $ \underset{u_i}{\mathrm{max}} \pi_i(u_i)$ is $\frac{1}{n_k}$,
thus $\eta \in [0, \frac{1}{n_k}]$.

However, in practice we clamp $ \pi^{\prime}_i(u_i^* | \overline{\varepsilon}) $
to allow a larger resonance probability $\eta \in [0, 1)$, 
at the cost of precision in individual policy invariance constraint.
Enabling this clamping,
we improve Eq.\ref{eq:ff1} into:
\begin{equation}
    \label{eq:ff2}
    \pi^{\prime}_i (u_i |  \overline{\varepsilon}) = \left\{\begin{array}{ll}
        \max\left[p_{\min},  \frac{\pi_i(u_i^*)-\eta}{1-\eta} \right], & u_i=u_i^* \\
        \pi_i(u_i )\left[\frac{1-     \pi^{\prime}_i \left(u_i^* |\overline{\varepsilon} \right)   }{1-\pi_i(u^*_i)}\right], & \text{otherwise}
    \end{array}\right.
\end{equation}
where $p_{\min} \le \frac{1}{n_k}$ is a small constant to avoid the situation $\pi^{\prime}_i(u_i^*) < 0$,
and we abbreviate $\max\left[p_{\min},  \frac{\pi_i(u_i^*)-\eta}{1-\eta} \right]$ 
to $\pi^{\prime}_i \left(u_i^* |\overline{\varepsilon} \right)$ when $u_i \neq u_i^*$.

A pseudo of Policy Resonance is presented in Appendix \ref{sec:Pseudo}.
Depending on whether the resonance state is allowed to change within an episode,
there are two options available for the code-level implementation, as shown in Fig.~\ref{fig:PolicyResonance}, 
We use the episode-level resonance by default 
and a comparison of these two variants is provided in Appendix.
In our practice, a two-stage training procedure is used.
During the first stage Policy Resonance is disabled ($\eta=0$).
Then in the second stage Policy Resonance is enabled,
with $\eta$ gradually increased from 0 to $\eta_{max}$.
When the training is finished (or halted for testing),
all procedures related to Policy Resonance will be removed,
and then agents start to make decisions according to raw individual policies $\boldsymbol{\pi}$
in a decentralized way (required by the CTDE paradigm).

%% file: 6-exp_results.tex
\section{Experiments}



This section demonstrates the effectiveness of the Policy Resonance approach.
Experiments are performed using AMD x86 CPUs and RTX8000 GPUs.

\subsection{Setup.}
\begin{itemize}
    \item We use episode-level resonance as default 
    (comparison with the step-level variant is in Appendix \ref{appendix:compare-Resonance-level}).
    \item Agents use shared policy network parameters to promote training efficiency, 
    which is a useful technique commonly used by Qmix, Conc4Hist and PPO-MA, etc.
    \item Applying a two-stage training procedure. 
    During the first stage the Policy Resonance (PR) is disabled, $\eta=0$.
    After $M$ episodes, the second stage begins with PR enabled, 
    increasing $\eta$ linearly from $0$ to $\eta_{max}$ in $M_{\text{pr}}$ episodes.
    \item We designed another variant model named \textbf{PR-Fast}, 
    which adopts a parameter-freezing trick to accelerate training.
    Specifically, 
    we split agent policy nerual network into body network and head network.
    (The head includes the logit-outputing layers in policy network).
    During the second stage in PR-Fast,
    body network parameters will be frozen to reduce computational cost.
    On the other hand, 
    the head network will be duplicated into $N$ copies for each agent, 
    allowing each individual to train their unique head network.
    \item The results are obtained from repeated experiments launched by different random seeds 
    in all figures except Fig.\ref{fig:PersonalityDevelop},
    which is a special figure illustrating some individual cases.
    \item \textbf{PR is disabled when evaluating data such as win-rate and test reward}.
\end{itemize}

\subsubsection{FME Experiments.} As described in Section \ref{BenchmarkEnvironments}, 
tags like \textit{1A4B2C} are used to describe the setting of an FME task.
E.g., \textit{1A4B2C} is a task composed of
one Lv-A, four Lv-B, and two Lv-C levels.
The maximum reward expectation in each level is 1,
thus the maximum reward expectation of \textit{1A4B2C} is 7.
The number of the agents $N$ varies between $[3,40]$,
and there are $n_k=10$ available actions in each level.
For both PR and PR-Fast implementation,
the first pre-training stage lasts for $M=2^{18} \approx 260k$ episodes
because FME is very computational cheap.
Additionally, $\eta_{max}=0.75$ and $M_{pr}=50k$.
Other related hyper-parameters are listed in Appendix \ref{appendix:alg}.

\subsubsection{ADCA Experiments.}
In ADCA, an RL-controlled team with $N=50$ agents is trained  
to compete against an opposite team (controlled by script AI) with more agents $N' \in [120\%N, 160\%N]$.
We use Conc4Hist as the host algorithm.
Training in ADCA is computationally expensive,
requiring a day to train the first stage $M=83k$.
In the second stage, the PR hyper-parameters are selected as 
$\eta_{max}=0.75$ and $M_{pr}=6400$ to complete the second stage rapidly.
Other related details is shown in Appendix \ref{appendix:adca} and \ref{appendix:alg}.

\subsection{Main Result.}
\begin{figure}[!t]  
    \vspace{-0.6cm}
    \centering
    \includegraphics[width=0.65\linewidth]{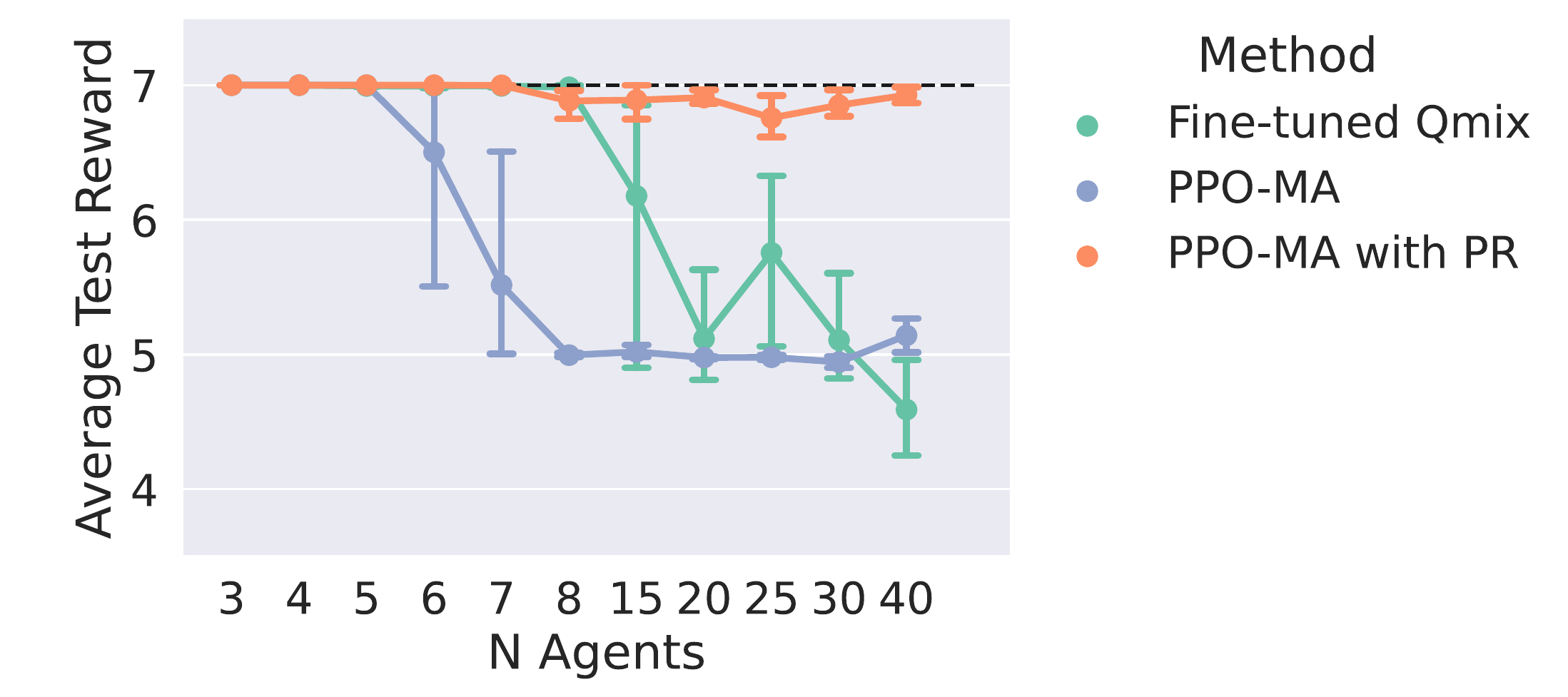}
    \vspace{-0.25cm}
    \caption{
        Evaluating PPO-MA, Fine-tuned Qmix and PPO-MA + PR
        in the FME diagnostic environment. FME setting is \textit{1A4B2C}.
        }
    \vspace{-0.5cm}
    \label{fig:DRWithMoreAgents}
\end{figure}

\begin{figure}[!t] 
    \centering
    \includegraphics[width=0.6\linewidth]{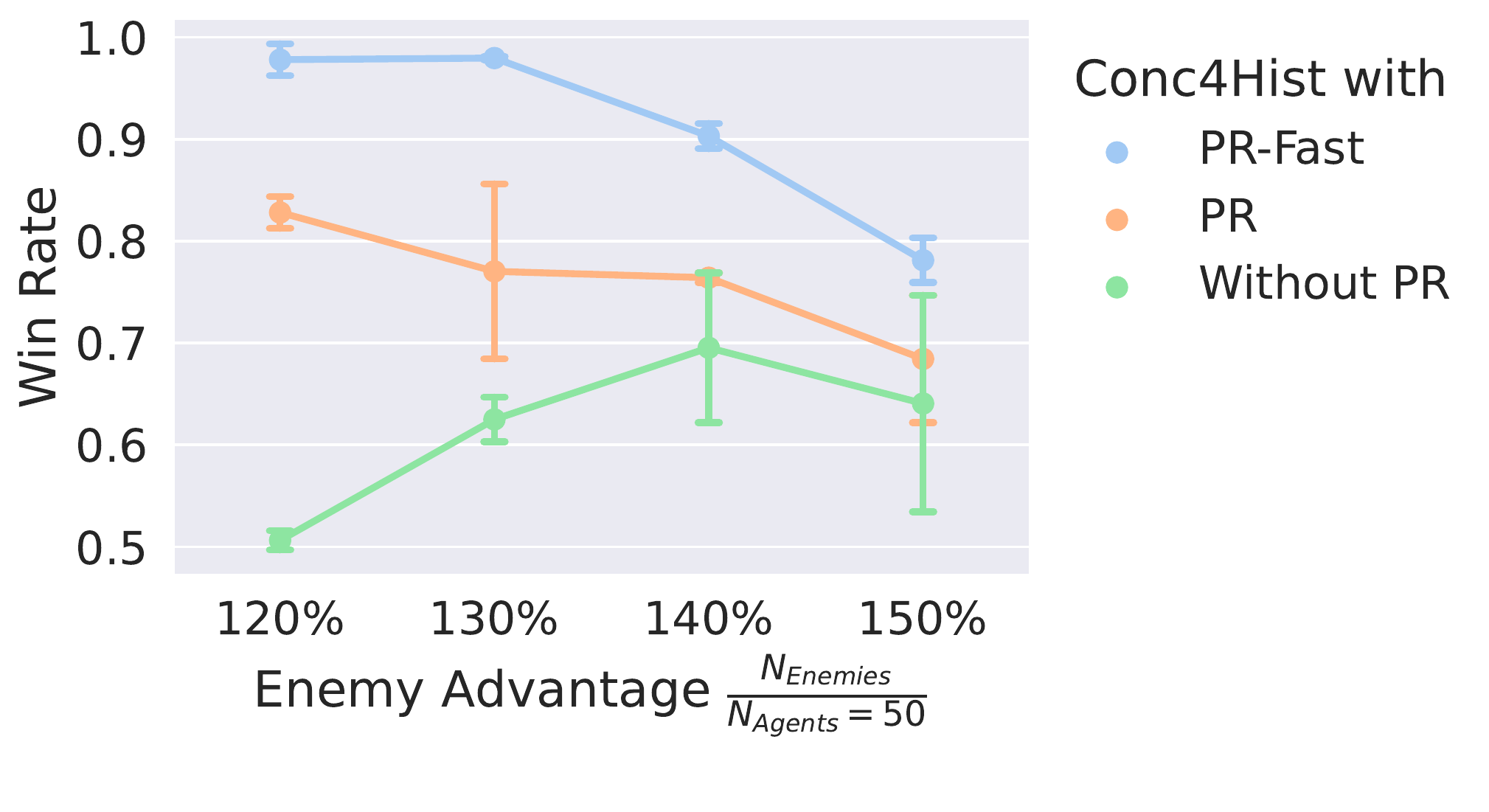}
    \vspace{-0.23cm}
    \caption{
        Win rates of algorithms tested in several ADCA settings.
    }
    \label{fig:ADCAWithDifferentMethods}
    \vspace{-0.5cm}
\end{figure}
\subsubsection{The Influence of Population Size $N$.}
The results of experiments on FME benchmark are displayed in Fig.\ref{fig:TwoStagePR-FME} and Fig.\ref{fig:DRWithMoreAgents}.
Note that the maximum reward expectation of \textit{1A4B2C} is 7, 
as is marked by dotted lines in these two figures.
It is apparent from Fig.\ref{fig:DRWithMoreAgents} that 
the DR problem causes a more significant impact as the number of agents grows larger.

Moreover, we can see that both Fine-tuned Qmix and PPO-MA suffer from the DR problem.
As a SOTA algorithm equipped with credit assignment (reward decomposition) techniques,
Fine-tuned Qmix can stand more agents than PPO-MA but still fails when $N \ge 15$.
\textbf{These results also indicate that credit assignment is not a solution to the DR problem}, 
because it does not change the exploration or exploitation behaviors of agents.

\subsubsection{The Effectiveness of Policy Resonance.}
Fig.\ref{fig:TwoStagePR-FME} and Fig.\ref{fig:TwoStagePR-ADCA} 
display the agents' learning process during the two-stage training process 
in FME benchmark and ADCA benchmark, respectively.
A reward breakthrough can be observed in both groups 
when PR is enabled at the beginning of the second stage.

Fig.\ref{fig:DRWithMoreAgents} and Fig.\ref{fig:ADCAWithDifferentMethods}
demonstrates the response of PR-enhanced algorithms towards 
the increase of agent population and task difficulty, respectively.
It can be seen in Fig.\ref{fig:DRWithMoreAgents} that PPO-MA implemented with PR
can easily outperform pure PPO-MA as well as Fine-tuned Qmix in FME.
Moreover, 
PPO-MA with PR can stably learn the optimal policy (dotted line in Fig.\ref{fig:DRWithMoreAgents})
even when $N \ge 30$.
For the ADCA benchmark,
Fig.\ref{fig:ADCAWithDifferentMethods} shows that 
Conc4Hist algorithm can promote the win rate from $50\%$ to above $80\%$ 
after implemented with PR under \textit{\footnotesize{120\%}-Enemy} setting.
In comparison, 
the performance of PR-Fast is slightly better.
PR-Fast can promotes \textit{\footnotesize{120\%}-Enemy} win-rate from $50\%$ to above $95\%$.
Even under very difficult \textit{\footnotesize{150\%}-Enemy} setting,
PR-Fast can still provide about 10\% win-rate improvement.

Furthermore, 
note that we do not need to identify DR problems before solving them.
We
\textbf{never assume or guess what form the DR problems might take in ADCA}.
Instead, 
we simply implement our PR approach on a proper host algorithm (Conc4Hist),
and then directly obtain a significant improvement.
This is because that PR is a general method and requires no task-specific knowledge.




\subsection{Ablations.}

\subsubsection{The Influence of $\eta$.}

\begin{figure}[!t]
    \vspace{-0.7cm}
    \begin{subfigure}{.49\linewidth}
      \centering
      \includegraphics[width=.76\linewidth]{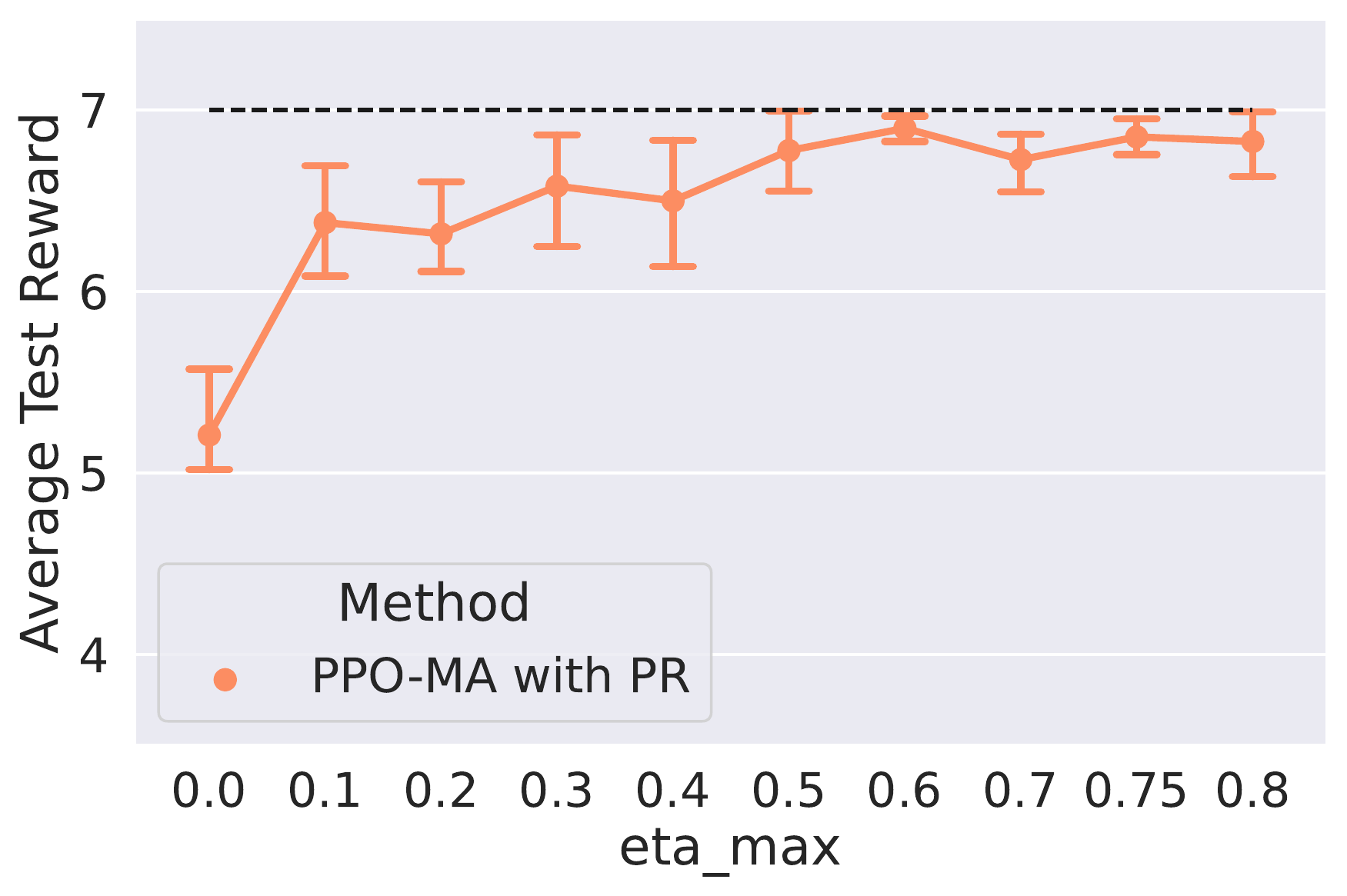} 
      \caption{
        FME. \textit{1A4B2C}, $N$=30.
      }
      \label{fig:abla-eta-FME}
      \label{fig:DRWithMoreAgents2}
    \end{subfigure}
    \begin{subfigure}{.49\linewidth}
      \centering
      \includegraphics[width=.75\linewidth]{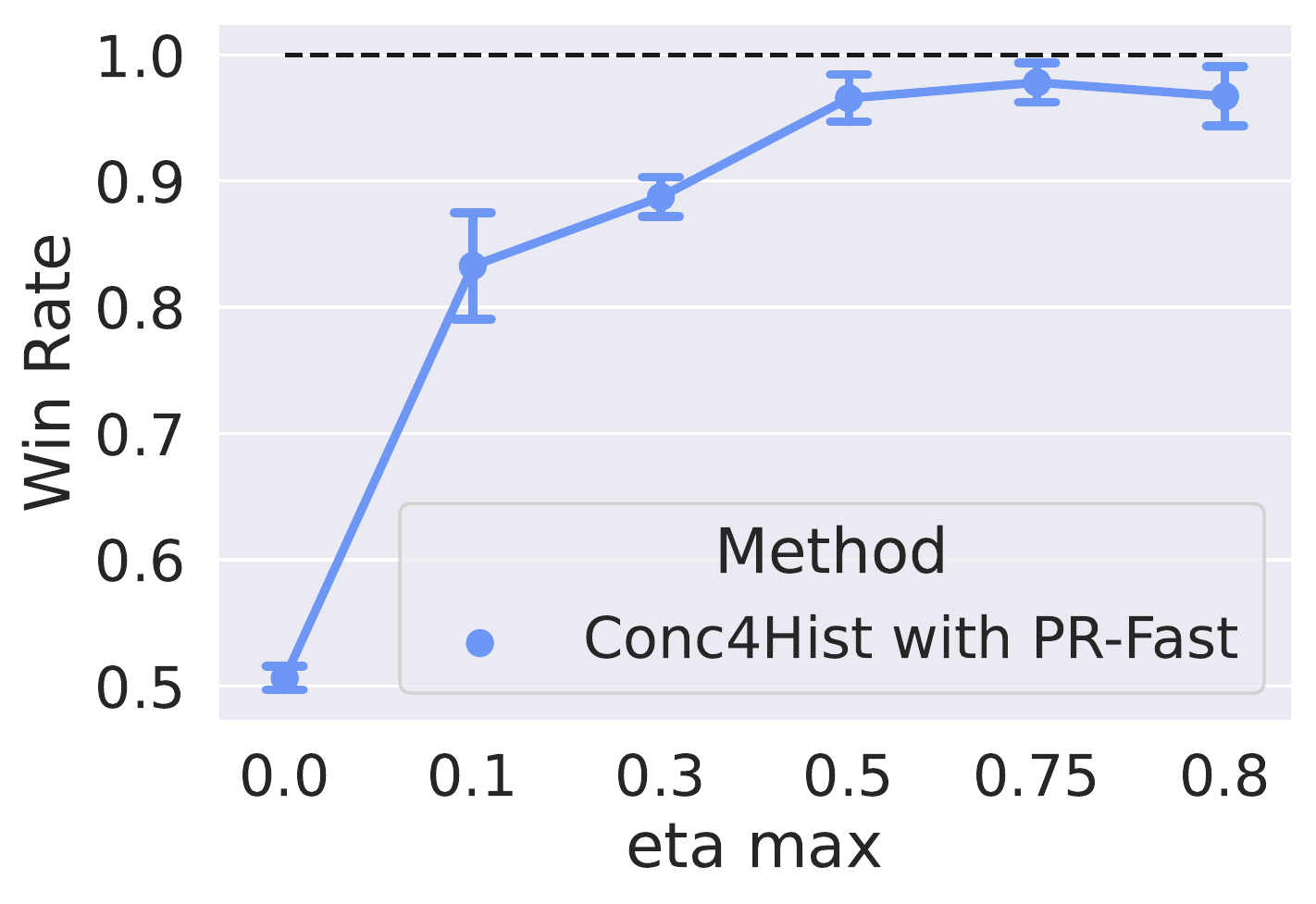} 
      \caption{
        ADCA. $N$=50, $N'$=120\%$N$.
      }
      \label{fig:abla-eta-ADCA}
      \label{fig:BasicMALevels}
      
    \end{subfigure}
    \vspace{-0.25cm}
    \caption{The influence of $\eta_{max}$.}
      \label{fig:influence-eta}
      \vspace{-0.5cm}
\end{figure}

The resonance probability $\eta$ is manipulated by $\eta_{max}$.
We run ablation tests to study the influence of $\eta_{max}$.
The results in Fig.\ref{fig:influence-eta} 
suggests that $\eta_{max} \approx 0.75$ is a good option.

\subsubsection{PR and PR-Fast.}
In FME, PR and PR-Fast show no significant difference (Fig.\ref{fig:TwoStagePR-FME}),
but in ADCA PR-Fast has advantage over PR (Fig.\ref{fig:ADCAWithDifferentMethods}).

\subsubsection{Step-Level PR and Episode-Level PR.}
The ablation study provided in Appendix \ref{appendix:compare-Resonance-level}
shows that they share almost identical performance.

%% file: 7-conclusion.md
This paper reports an MARL problem named the Diffusion of Responsibility (DR), 
which has a negative impact in
both policy-based and value-based multiagent reinforcement learning (MARL) algorithms.
Firstly,
we analyze the feature and the cause of the DR effect.
And we reveal that the DR problem can be traced back to the
flaws in current multiagent exploration-exploitation dilemma solutions in existing MARL algorithms.
Secondly, a Policy Resonance method is proposed to solve the DR problem.
To demonstrate the effectiveness of our Policy Resonance solution,
we run experiments with multiple distinct MARL algorithms on multiple MARL benchmarks, 
including FME and ADCA tasks.
The results show that our Policy Resonance approach is effective against DR problems.

%% file: 8-apendix_real.tex
\clearpage

\appendix
\section{Appendix}
\subsection{Lv-B Optimal Reward Expectation.} \label{lvb-reward}

In Lv-B the team is rewarded according to number of agents selecting arm $u^\beta$, 
namely $R_B=N_{\mathcal{A}}(u^\beta)/N*c_\beta$.
Here $N_{\mathcal{A}}(u^\beta)$ counts the number of agents selecting $u^\beta$, 
$c_\beta$ is a constant making the optimal reward expectation fixed to 1. 
Let $u^{\beta*}= \underset{u}{\mathrm{argmax}}  [p_\beta(u)]$,
then $c_\beta=1/{\underset{u}{\mathrm{max}}[p_\beta(u)]}=1/ {p_\beta(u^{\beta*})}$.
Let $\boldsymbol{\pi} = \lbrace \pi_1,\pi_2, \dots \rbrace$ be the joint policy,
then the reward expectation is:
\begin{equation}
    \label{eq:typeb-expect}
    \begin{aligned}
    \mathbb{E}\left[R_B|\boldsymbol{\pi}\right] 
    &=\Sigma_{i} \Sigma_{k} {\pi_{i}\left(u^{k}\right) 
    p_{\beta}\left(u^{k}\right)} c_{\beta}/{N} \\
    &=c_{\beta} \Sigma_{k} p_{\beta}\left(u^{k}\right)  {\Sigma_{i} 
    \pi_{i}\left(u^{k}\right)}/{N}  \\
    &\leq c_{\beta} p_{\beta}\left(u^{\beta*}\right) 
     {\Sigma_{k} \Sigma_{i} \pi_{i}\left(u^{k}\right)}/{N} \\
    & \leq 1
    \end{aligned}
    \end{equation}
where $\pi_{i}\left(u^{k}\right)$ represents the probability of $i$-th agent choosing action $u^k \in \mathcal{U}$.
The optimal solution is all agents chosing $u^{\beta*}$, which is the condition for equality in Eq.\ref{eq:typeb-expect}.

\subsection{Details of FME.}\label{appendix:FMEtable}
In this paper, 
the FME diagnostic environment is used to investigate the DR problem.
Each FME task is composed of a series of levels,
and at each level, agents need to make single-step decisions before entering the next one.
The team reward obtained at a level is immediately given to RL algorithms for policy optimization.

\begin{figure}[!t] 

	\centering
	\includegraphics[width=\linewidth]{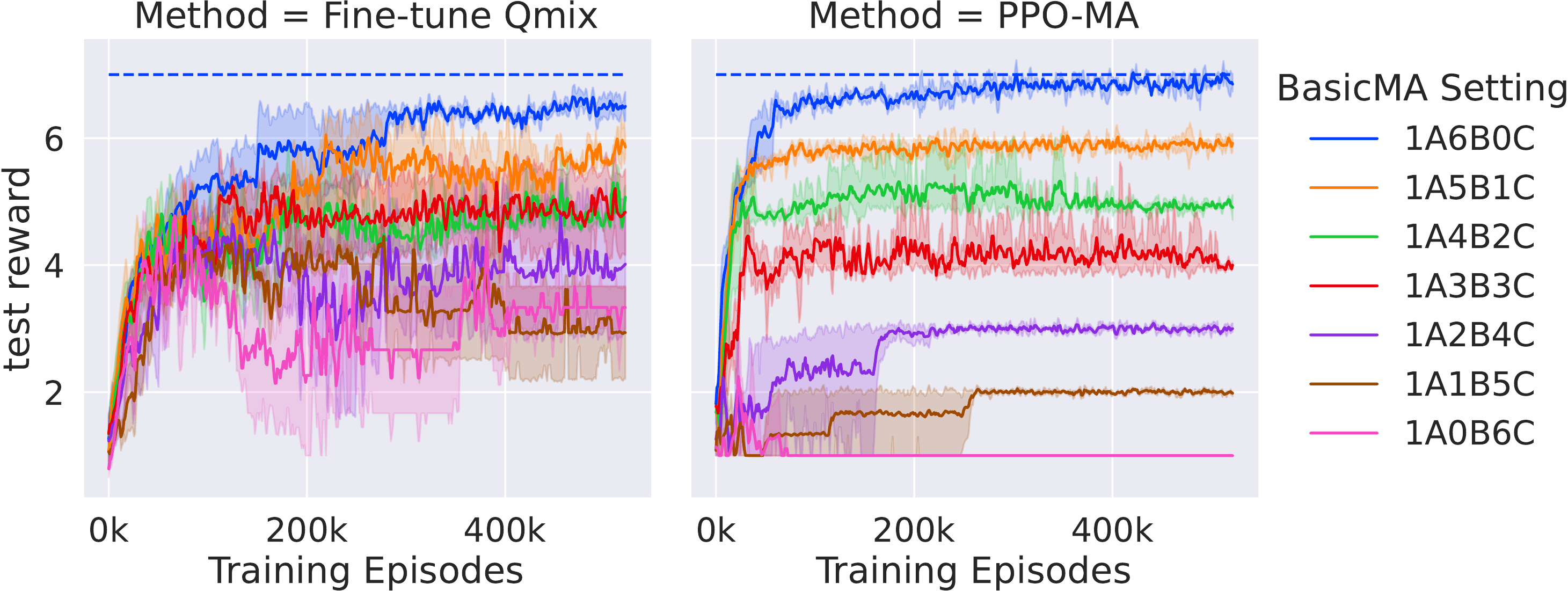}
	\caption{The comparison of policy-based algorithms (PPO-MA) and value-based algorithms (Fine-tuned Qmix) under several FME settings.
	}
	\label{fig:FMELevels}

\end{figure}

As described previously, there are three types of levels in FME.
One type of level can be reused multiple time in a task,
as we can make them different by changing the configuration.
E.g., in the first Lv-B level of our \textit{1A2B4C} task,
the actions with top probability to be rewarded is: 
\begin{itemize}
    \item Lv-B configuration \textit{Act-5-6-7}, which indicates that the probabilities of rewarding
	\textit{Act-5}, \textit{Act-6}, \textit{Act-7} are 50\%, 40\%, 10\% respectively.
\end{itemize}
but the second Lv-B level has an alternative configuration: 
\begin{itemize}
    \item Lv-B configuration \textit{Act-6-7-8}, which indicates that the probabilities of rewarding 
	\textit{Act-6}, \textit{Act-7}, \textit{Act-8} are 50\%, 40\%, 10\% respectively.
\end{itemize}
The probability descent pattern is fixed to $[50\%, 40\%, 10\%]$ in all Lv-B configurations.
Configurations of all FME settings used in experiments are listed in Tabel.~\ref{basicma_configurations}.
Fig.\ref{fig:FMELevels} shows an additional experiment 
which tests Fine-tuned Qmix and PPO-MA under different FME settings.

\begin{table*}[!h]

	\centering
	\setlength{\tabcolsep}{1.5mm}{
		\begin{tabular}{cccccccc}
		\toprule
	 & \textbf{1A6B0C}                           
	 & \textbf{1A5B1C}                           
	 & \textbf{1A4B2C}                           
	 & \textbf{1A3B3C}                           
	 & \textbf{1A2B4C}                           
	 & \textbf{1A1B5C}                           
	 & \textbf{1A0B6C}  \\
		\midrule
		Lv.1   & Act-0                             & Act-0                             & Act-0                             & Act-0                             & Act-0                             & Act-0                             & Act-0    \\
		Lv.2   & Act-4-5-6 & Act-4-5-6 & Act-5-6-7 & Act-5-6-7 & Act-5-6-7 & Act-5-6-7 & Act-0    \\
		Lv.3   & Act-5-6-7 & Act-5-6-7 & Act-6-7-8 & Act-6-7-8 & Act-6-7-8 & Act-0                             & Act-1    \\
		Lv.4   & Act-6-7-8 & Act-6-7-8 & Act-7-8-9 & Act-7-8-9 & Act-0                             & Act-1                             & Act-2    \\
		Lv.5   & Act-7-8-9 & Act-7-8-9 & Act-8-9-5 & Act-0                             & Act-1                             & Act-2                             & Act-3    \\
		Lv.6   & Act-8-9-0 & Act-8-9-5 & Act-1                             & Act-1                             & Act-2                             & Act-3                             & Act-4    \\
		Lv.7   & Act-9-0-1 & Act-0                             & Act-0                             & Act-2                             & Act-3                             & Act-4                             & Act-5   \\
		\midrule
		$\mathbb{E}\left[R|\boldsymbol{\pi}^*\right]$  & 7.0 & 7.0  & 7.0 & 7.0 & 7.0    & 7.0      & 7.0   \\
		\bottomrule
	\end{tabular}}
	\caption{
        The level configurations of each tagged FME task are used in this paper ($n_k = 10$).
        For Lv-A and Lv-C levels, this table gives the selection of the $u^\alpha$ and $u^\gamma$, e.g. \textit{Act-0}.
        For Lv-B levels, a action sequence is given, 
        e.g., \textit{Act-4-5-6} indicate that the probability of rewarding \textit{Act-4}, \textit{Act-5}, \textit{Act-6} is 50\%, 40\%, 10\% respectively.
        The probability descent pattern is fixed to $[50\%, 40\%, 10\%]$ in all Lv-B configurations.
        The reward expectation of an optimal learned policy $\mathbb{E}\left[R|\boldsymbol{\pi}^*\right]$ 
        equals to the number of levels in these tasks.
	}
	\label{basicma_configurations}
	\end{table*}


\subsection{Details of ADCA.}\label{appendix:adca}

The Asymmetric Decentralized Collective Assault (ADCA)
benchmark is a swarm combat environment modified and 
improved from \cite{fu2022concentration}.
As shown in Fig.\ref{fig:TwoStagePR-ADCA}, in ADCA, 
two asymmetric teams compete with each other,
the RL-controlled team has $N=50$ agents and suffers from
limited observation as well as observation interference.
On the other hand, 
the opposite team that is controlled by script AI
has much more agents to use ($N' \in [120\%N, 150\%N]$),
observes all the field information for making decisions
and has faster weapon targeting speed.

All agents control their movement by adjusting the direction 
of their acceleration.
Discrete actions available to agents are responsible
for selecting acceleration and rotating weapon directions.
To address the challenge of outnumbered situation and jammed observation,
the team needs to learn to strike back by taking terrain advantages,
which slightly increases the effective weapon range.
Agents are rewarded according to their kill-death ratio.
The winning condition is to eliminate all opponents,
or at least to reverse the disadvantage in agent number 
when time is up.

\begin{figure*}[!t]
	\centering
	\includegraphics[width=\linewidth]{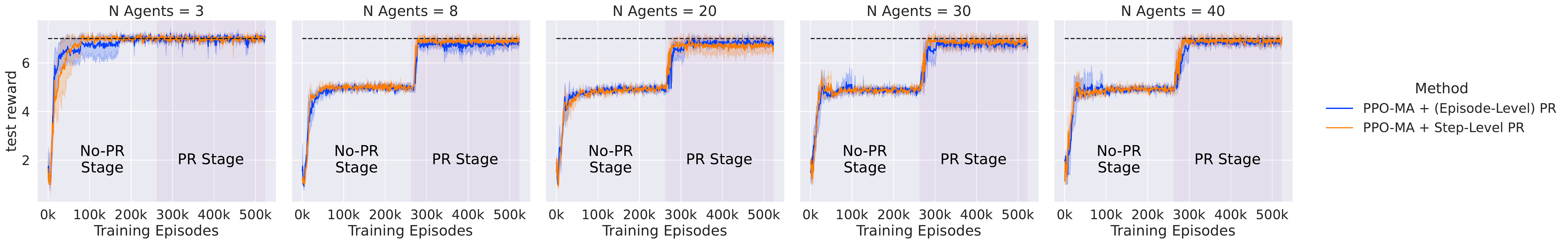}
	\caption{
        Comparing episode-level and step-level Policy Resonance in FME.
        The setting is \textit{1A4B2C}, $N=30$ and $n_k=10$.
	}
	\label{fig:EpisodeLevelStepLevelCompare}
\end{figure*}

\subsection{DR Problem: Value-Based Perspective.} \label{Value-Based-Perspective}
The analysis of the DR effect in value-based Methods is easier.
The exploration of most value-based RL methods is powered by $\epsilon$-greedy 
for carrying out exploration.
To guarantee that agents can always discover other possible policies better than the current ones,
agents always have a small probability of acting against the advice of value functions (Q functions).
Suppose there are $n_k$ discrete actions in FME Lv-C level, 
and all agents mistakenly value the main action $u^\gamma$ more than other actions,
then the probability for agents to notice this mistake in an episode is:
\begin{equation}
    P_{penalty} = \left[(1 - \epsilon) + \frac{\epsilon}{n_k} \right]^N 
\end{equation}
where $(1 - \epsilon) + \frac{\epsilon}{n_k} < 1$,
and $P_{penalty}$ will decrease exponentially with the $N$ becomes larger.
When the agent population is large enough,
the agents will mistakely treat ALL-$u^\gamma$ policy as a sub-optimal solution
to finish the task,
with each individual expecting some other agents to take the responsibility
to carry out non-main actions.
This theory is backed by experiment results shown in Fig.~\ref{fig:BasicMALevels},
where the Fine-tuned Qmix \cite{rashid2018qmix, hu2021rethinking}, a well-developed value decomposition method,
is used to solve the FME tasks.
Qmix fails when $N$ is large and numerous Lv-C levels are present.

\input{12apendix_code.tex}

\subsection{MARL Algorithms Used in Experiments.}\label{appendix:alg}
\input{h-table.tex}

PPO-MA is an algorithm that resembles much of the MAPPO model proposed in \cite{yu2021surprising}.
However, 
most tricks used in \cite{yu2021surprising} are not adopted to keep the model as simple as possible, e.g., the trick of action masking and input shaping.
Moreover, we used dual-clip PPO \cite{ye2020mastering} instead of the classic version
to increase training stability.

Conc4Hist is a model based on the Concentration Network proposed in \cite{fu2022concentration}.
This model is efficient in dealing with large-scale multiagent problems and
can consider multi-step observation to work against interference 
that may exist in sophisticated MARL environments.

Fine-tuned Qmix is a value decomposition algorithm specialized in solving 
small-scale cooperative multiagent tasks such as SMAC.
In \cite{hu2021rethinking} the authors discovered that 
code-level implementation tricks significantly influence the performance of Qmix.
After tuning, Qmix surprisingly outperforms 
many SOTA methods in the value-decomposition MARL algorithm family.

There are significant differences between these algorithms,
First and most importantly,
the former two algorithms are policy-based, but Qmix is value-based.
Secondly, 
Qmix should be categorized into off-policy models while 
the former two algorithms are on-policy models considering 
how they utilize the collected samples.
Thirdly,
Qmix is designed with the function of breaking down team rewards
into individual rewards,
while PPO-MA and Conc4Hist simply dispense team rewards equally to each agent.
When individual rewards are not available, e.g., in FME.
The details of the hyper-parameters for each algorithm are shown in Table.~\ref{hyper-para}.

\subsection{Ablations of the Resonance Probability.}
As the core hyper-parameter in Policy Resonance,
the resonance probability $\eta$ is our primary concern during Policy Resonance.
We carry out an ablation study of $\eta$ by changing the maximum $\eta$ limit $\eta_{max}$.
Recall that $\eta$ increase from 0 to $\eta_{max}$ in $M_{pr}$ episodes,
then maintains at $\eta_{max}$ till the end of training.
We set the $M_{pr}$ value identical to previous experiments,
and the impact of shifting $\eta_{max}$ is illustrated in 
Fig.~\ref{fig:DRWithMoreAgents2} and Fig.~\ref{fig:BasicMALevels}.

In both FME and ADCA,
a great leap in performance is observed even when the $\eta_{max}$ is as low as $0.1$.
Nevertheless, a larger $\eta_{max}$ setting around $0.7$ 
reduces the test reward variance (when experimented repeatedly) in FME,
and benefits the winning rate in ADCA.

\subsection{Ablations of PR and PR-Fast.}
In the Policy Resonance stage, it is not necessary to train the whole policy neural network,
instead, 
freezing the shared parameters that are not processed by the policy head networks
can accelerate the learning process.
To verify the influence of the two method variants,
we provide a comparison study in both PPO-MA and ADCA tasks.

Fig.~\ref{fig:TwoStagePR-FME} and \ref{fig:ADCAWithDifferentMethods} demonstrate the comparison of 
the performance differences depending on whether or not to freeze the shared parameters.
In PPO-MA tasks,
the difference between the two method variants is insignificant.
In contrast,
this parameter freezing technique plays a vital role in promoting the winning rate in ADCA.

%% file: 12apendix_code.tex
\subsection{Pseudo.}\label{sec:Pseudo}

\begin{algorithm}[h]
    \caption{Episode-Level Policy Resonance}
    \label{alg:Policy_Resonance}
    \begin{algorithmic}[1] 
    \STATE Initialize resonance probability $\eta=0$.
    \STATE Wait Policy Resonance to be enabled.
    \FOR {episode $M_e = 1,2,\dots, M_{pr}$}
        \STATE $\eta = \eta_{max} *  M_e/M_{pr} $
        \STATE \textcolor{blue}{Randomly choose resonated state between $\varepsilon$ and $\overline{\varepsilon}$ with probability $P(\varepsilon)=\eta$.}
        \FOR {step $t = 1,2,\dots$}
            \STATE \textcolor{blue}{{(* Put Line 5 here for Step-Level Policy Resonance.)}}
            \IF {$\varepsilon$}
                \STATE Each agent acts according to Eq.\ref{resonatedindividual}.
            \ELSIF {$\overline{\varepsilon}$}
                \STATE Each agent acts according to Eq.\ref{eq:non-resonated} and \ref{eq:ff2}, with the clamping enabled.
            \ENDIF
        \ENDFOR
    \ENDFOR
    \STATE Continue training with $\eta = \eta_{max}$, until the training is complete.
    \end{algorithmic}
    \end{algorithm}



\subsection{Step-Level and Episode-Level Policy Resonance.}\label{appendix:compare-Resonance-level}

Previously we proposed two variants of Policy Resonance, namely step-level and episode-level Policy Resonance, 
which have a tiny difference in code implementation, as shown in Alg.\ref{alg:Policy_Resonance} and Fig.~\ref{fig:PolicyResonance}.

In a ablation study shown in Fig.~\ref{fig:EpisodeLevelStepLevelCompare},
two variants have similar performance. 
Nevertheless, we still choose the episode-level resonance as the default solution. 
Considering that it may take multiple time steps for an agent to finish a certain behavior, 
changing the resonance state within an episode may break up agent behaviors and leads to negative results.

%% file: h-table.tex
\begin{table*}
	\caption{The hyper-parameters of algorithms used in this paper.}
	\label{hyper-para}
	\centering
	\setlength{\tabcolsep}{1.5mm}{
		\begin{tabular}{c c c c}
		\toprule
			  &  \textbf{PPO-MA} 	& \textbf{Conc4Hist} & \textbf{Fine-tuned Qmix} \\
		\midrule
			learning rate 		&     	
			$5\times 10^{-4}$  	&
			$5\times 10^{-4}$  	&
			$1\times 10^{-3}$  \\
			on-policy batch size (n episodes)	&      		
				512	&	   
				64   &         
				-       \\
			off-policy batch size (n episodes)	&      		
				-	&	   
				-   &         
				64       \\
			off-policy memory storage (n episodes)  	&      		
				-	&	   
				-   &         
				5000       \\
		special neural networks utilization 		&     	
			-  	&
			ConcNet  	&
			Mixing Network  \\
			BP n times each update  &      
			24  	&     
			24      &        
			1      \\
			epsilon anneal step		&     	
			-  	&
			-  	&
			200000  \\
			optimizer 		&     	
			Adam  	&
			Adam  	&
			Adam  \\
			gamma 		&     	
			0.99  	&
			0.99  	&
			0.99  \\
		
			min epsilon		&     	
			-  	&
			-  	&
			0.1  \\
			PPO clamp		&     	
			0.2  	&
			0.2  	&
			-  \\
			GAE $\lambda$ or TD $\lambda$ 		&     	
			0.95  	&
			0.95  	&
			0.6  \\

			entropy	coefficient 	&     	
			0.05  	&
			0.2  	&
			-  \\
			averge n episodes in each test	&     	
			192  	&
			64  	&
			192  \\
			hidden layer typical dimension 	&     	
			8  	&
			48  &
			32  \\
			update target network every n episodes	&     	
			- 	&
			-  &
			200  \\
			concentration pruning preserve $d_c$	&     	
			- 	&
			3 &
			-  \\
		\bottomrule
	\end{tabular}	}

\end{table*}